\documentclass[11pt, twocolumn]{article}

\usepackage[letterpaper, left=2.5cm, right=2.5cm, bottom=2cm, top=2cm]{geometry}

\usepackage{graphicx}
\usepackage{amsfonts}
\usepackage{amsmath}
\usepackage{booktabs}
\usepackage{float}
\usepackage{hyperref}
\usepackage{multicol}

\usepackage{adjustbox}

\AtBeginDocument{%
\providecommand\BibTeX{{%
\normalfont B\kern-0.5em{\scshape i\kern-0.25em b}\kern-0.8em\TeX}}}

\begin{document}
\pagenumbering{gobble}


\title{\textbf{Electricity Theft Detection with self-attention}}

\author{
	\texttt{Paulo Finardi}$^1$\footnote{Both authors contributed equally to this research}   
	\and \texttt{Israel Campiotti}$^1$ \footnotemark[\value{footnote}]  
	\and \texttt{Gustavo Plensack}$^1$   
	\and \texttt{Rafael Derradi de Souza}$^1$   
	\and \texttt{Rodrigo Nogueira}$^2$   
	\and \texttt{Gustavo Pinheiro}$^3$   
	\and \texttt{Roberto Lotufo}$^1$  \\ \\
	\small{\{paulo.finardi, israelcampiotti, gustavo.plensack, rafael.souza, rodrigo.nogueira, roberto\}@neuralmind.ai}\\
	\small{gustavo@onerf.com.br} \\ \\
        \small{$^1$NeuralMind, S\~ao Paulo, Brazil $\;$  $^2$NeuralMind, New York, USA $\;$ $^3$OneRF Networks, S\~ao Paulo, Brazil}
}

\date{}

\maketitle

\begin{abstract}
\textit{In this work we propose a novel self-attention mechanism model to address electricity theft detection on an imbalanced realistic dataset that presents a daily electricity consumption provided by State Grid Corporation of China. Our key contribution is the introduction of a multi-head self-attention mechanism concatenated with dilated convolutions and unified by a convolution of kernel size $1$. Moreover, we introduce a binary input channel (Binary Mask) to identify the position of the missing values, allowing the network to learn how to deal with these values. Our model  achieves an AUC of $0.926$ which is an improvement in more than $17\%$ with respect to previous baseline work. The code is available on GitHub at \href{https://github.com/neuralmind-ai/electricity-theft-detection-with-self-attention}{github.com/neuralmind-ai/electricity-theft-detection-with-self-attention}.}
\end{abstract}


\section{Introduction}

According to the World Bank, in $2017$ more than $88\%$ of the world population had access to electrical energy, which is made available to people via a complex transmission and distribution system that interconnects power plants to consumers.
In the operation of this system two types of losses are expected: technical and non-technical losses. 
Technical Losses (TL) occur due to power dissipation in the materials that compose the electrical power system itself, such as cables, connectors, and power transformers.
Non-Technical Losses (NTL) represent energy losses due to energy theft and errors of billing or measurement \cite{Glauner_2017}.

According to the Electricity Distribution Loss Report published by ANEEL (Brazilian National Electricity Agency) \cite{Aneel2019}, NTLs comprised about $6.6\%$ of all energy injected into the Brazilian electrical power system in $2018$.
These losses impact consumers with more expensive energy bills, electricity distribution companies with reduced revenues, and the reliability of the electrical power system \cite{Chaunan2013}.
Part of the of the problem of tackling NTLs is dealing with the metering infrastructure itself, which is pointed out as being the most faulty subsystem \cite{Chaunan2013}.
Recent advances in the Internet of Things (IoT) made possible addressing these problems by the adoption of Advanced Metering Infrastructures (AMIs), that can provide consumption data with high temporal resolution, thus reducing losses related to billing and metering issues.
Together with AMIs, artificial intelligence algorithms can play an important role in detecting NTLs due to electricity theft in power distribution system \cite{liu2019,Zheng2018}.

In this work, we developed a predictive method using a supervised learning technique with deep learning methodologies applied to to identify fraudulent consumer units. We train and evaluate our models on a dataset of $34$ months of daily electricity consumption.
The work brings several improvements compared with the previous state-of-the-art method \cite{Zheng2018}, such as the usage of Quantile normalization on the original data, the usage of an additional binary input channel to deal with missing values and the usage of attention mechanism.

Our results show that the usage of a model with attention mechanism layers delivered an increment of $17\%$ on the Area Under the Curve (AUC) score when compared to the baseline.
The combination of this model with a the binary input channel (Binary Mask) and Quantile normalization improved the AUC and the F$_1$.  

The article is organized as follows: in section \ref{sec:related_work} we present an overview of related works; in section \ref{sec:problem_analysis} we present the problem and the methodology adopted; in section \ref{sec:architecture_overview} we detail the proposed solution and the metrics used to evaluate the performance of the algorithms; in section \ref{section_data} we describe the data processing steps; section \ref{sec:experiments} presents the results obtained; and finally, in section \ref{sec:conclusion} we describe our conclusions and future perspectives.

\section{Related Work}
\label{sec:related_work}

The application of deep learning in NTLs detection has increased in recent years. Several approaches to the problem have been proposed and the results obtained are significantly superior when compared to those from rule-based traditional methods \cite{Glauner_2017,li_multi-scale_2018,Zheng2018}. However, one of the main difficulties in developing data-driven models for NTLs detection in the electricity industry is the lack of data publicly available.
Energy consumption is a sensitive data and due to privacy and security issues the vast majority of electricity distribution companies do not share their data. One of the ways to circumvent this problem is to generate synthetic data.
For instance, Liu et al.~\cite{liu2019} inject artificial electricity thefts into a database of regular consumers.
Although useful, the generation of synthetic data may lead to unintentional introduction of bias or misrepresentation of real situations.

Zheng et at.~\cite{Zheng2018} present a study using a dataset with real electricity theft data provided by State Grid Corporation of China (SGCC). This study, which has become a baseline for following recent works, introduces a neural network architecture based on a wide (dense) and a deep (convolutional) component trained together. Moreover, their proposed reshaping of the 1D electricity consumption data sequences into a 2D format has provided a straightforward way to explore neighborhood correlations with 2D convolutional neural network (CNN).
Hasan et al.~\cite{hasan2019} uses real electricity theft data, they propose a combination of CNN and LSTM (Long Short-Term Memory) architectures in order to explore the time-series nature of the electricity consumption data. Nonetheless, satisfactory results were achieved only after applying the synthetic minority over-sampling technique (SMOTE) \cite{Yang_2016} to account for the imbalanced dataset.

In Li et al.~\cite{li_electricity_2019}, a combination of CNN with Random Forest (RF) algorithm is applied on a dataset of over 5000 residential and businesses consumers provided by the Electric Ireland and Sustainable Energy Authority of Ireland (SEAI), with thieves being synthetically injected. Also motivated by the data reshaping introduced by Zheng et al. (2018), the authors reshaped the electricity consumption data into a 2D format, allowing a more generalized feature extraction by the CNN.

\section{Problem Analysis}
\label{sec:problem_analysis}
Our task is to detect fraud in electricity consumption. The dataset is a collection of real electricity consumption samples and was released by the State Grid Corporation of China (SGCC). 
The data is a sequence of daily electricity consumption, which we characterize as a time series. The basic assumption that guides the analysis of time series is that there is a causal system more or less constant, related to time, which influenced the data in the past and can continue to do so in the future. The purpose of time series analysis is to identify nonrandom patterns in the daily electricity consumption behavior that allows more accurate predictions. See section \ref{section_data} for a time series analysis and autocorrelations for the problem at hand.

\subsection{Data Methodology}
\begin{figure}
\centering
	\caption{\small{\textbf{Data processing methodology.}}}
\includegraphics[width=8cm, height=8cm]{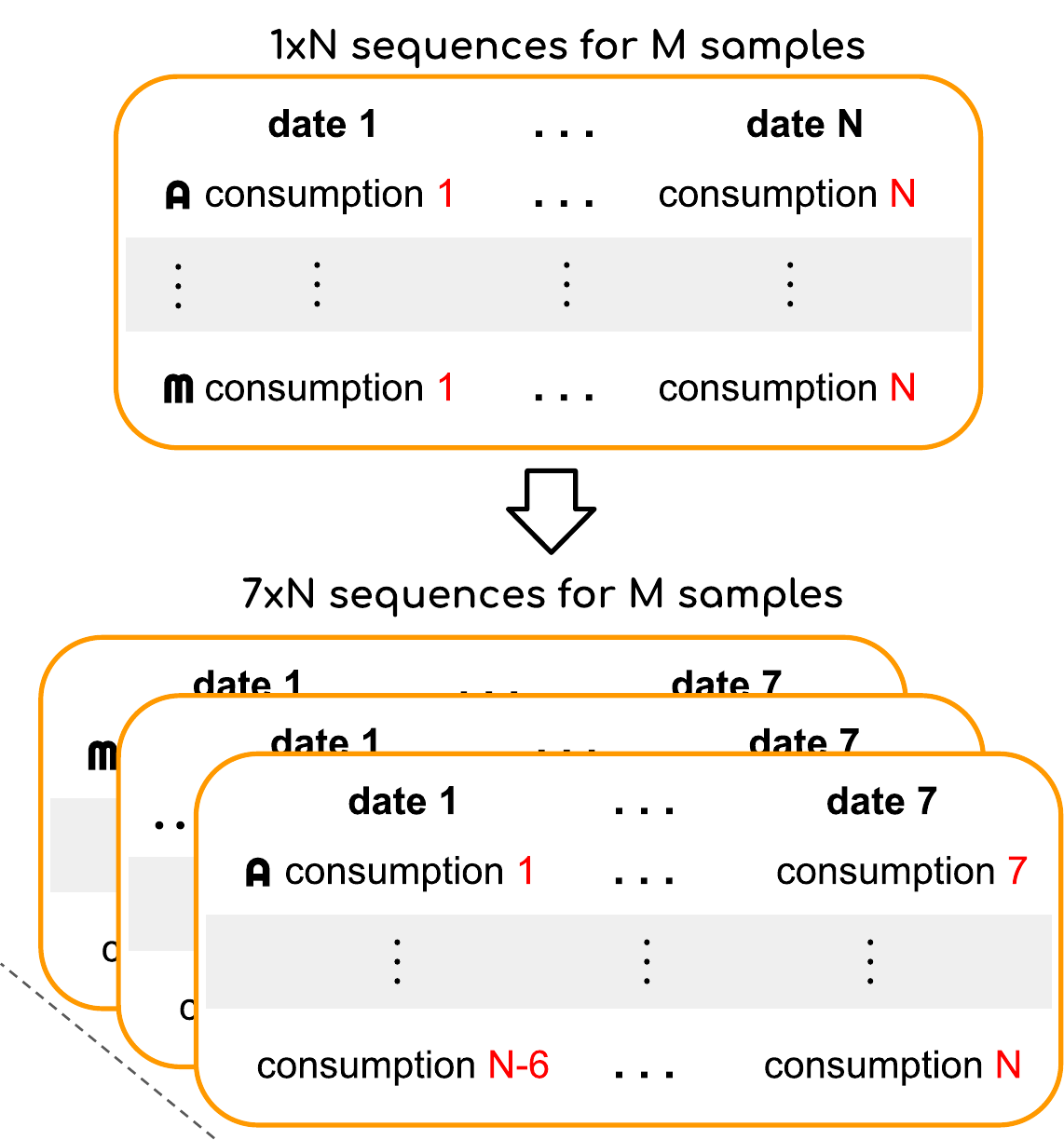} 
\label{data_methodology}
\end{figure}

A important contribution from Zheng et al.~\cite{Zheng2018} is the transformation of one dimensional data into bidimensional (Figure \ref{data_methodology}). A 2D format allows the exploration of periodicity and neighborhood characteristics with the usage of a computer vision models, such as 2D convolutional neural networks. 

\subsection{Missing data}
\label{section_missing_data}
Missing data is an ubiquitous problem. In the literature we find two common practices to deal with them. One approach is to delete the incomplete reading from the dataset. However, this approach may dismiss valuable information. An alternative is to estimate the missing value using interpolation or with the median or mean of the data feature \cite{nan-tradictional}. Although those techniques have been proven effective, they impose strong assumptions about the nature of the missing data and hence might bias the predictive models negatively.
In addition to these methods, attempts to find approximations to fill the missing data using genetic algorithm, simulated annealing and particle swarm optimization have also been proposed \cite{nan-optim}. However, when dealing with large datasets such techniques can be prohibitively slow.

\begin{figure}
\centering
	\caption{\small{\textbf{Top left: raw data in 2D format, Top right: missing entries are filled with 0's, Left bottom: binary mask, Right bottom: final data with 2 channels. 
	\vspace{0.222cm}}}}
\includegraphics[width=6cm, height=6cm]{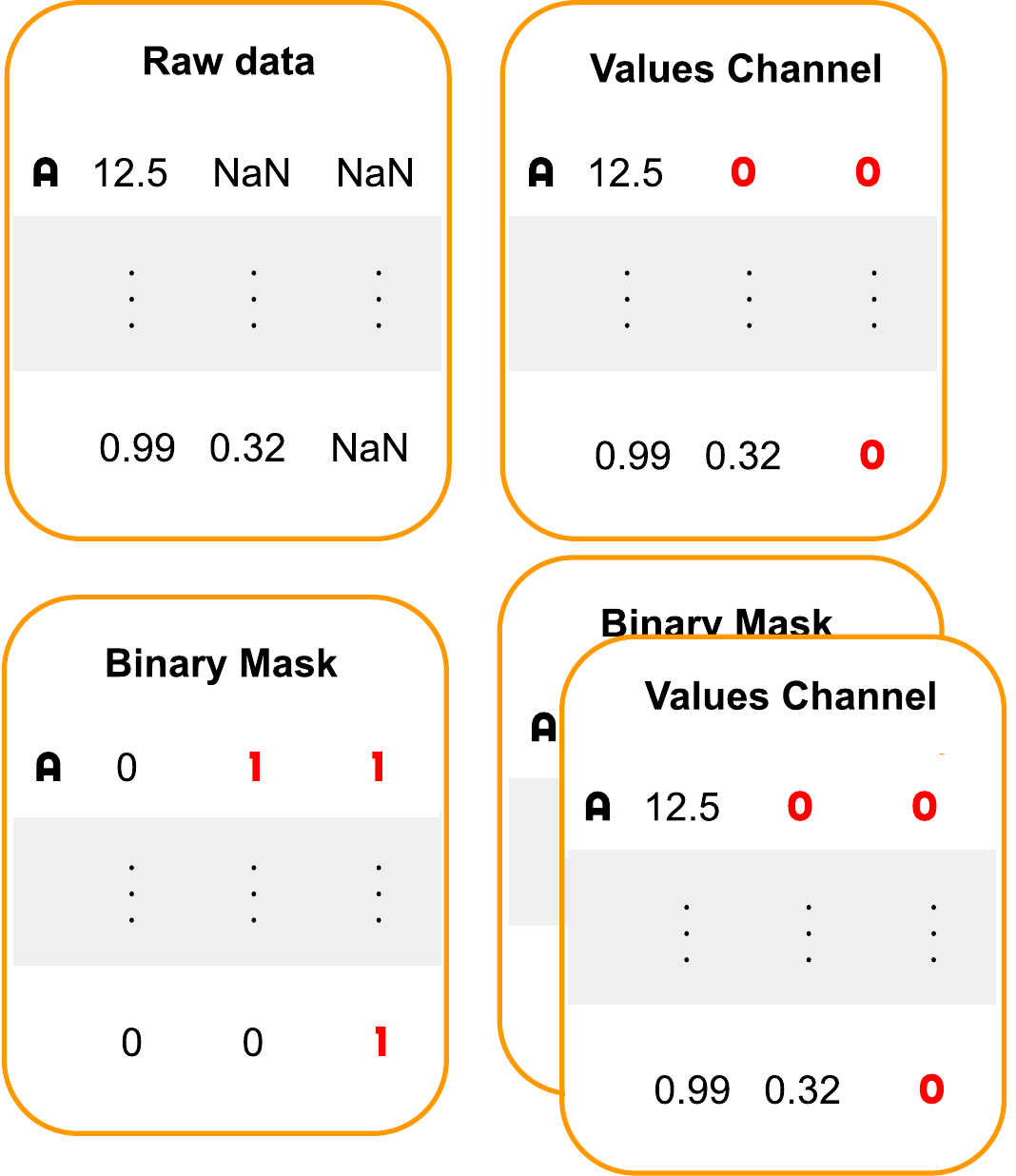} 
\label{figure_nan}
\end{figure}

To deal with the missing values, we create a binary mask as an additional channel of the input as follows: First, we identify the indices of all missing data and create a binary mask, where the missing data receives value $1$ and all remainder values receives $0$. We call this mask \textit{Binary Mask}. The missing data at the \textit{values} channel receives a value of $0$. These $2$ channels are the input to a 2D CNN. See Figure ~\ref{figure_nan} for an illustration of our method.

\section{Architecture overview}
\label{sec:architecture_overview}
Image recognition is a classic classification problem where CNNs have a history of high efficacy~\cite{lecun98, alexnet}. As our data input resembles an image, we developed two models to address the problem, both using 2D convolutions: a CNN and a multi-head attention model. Attention models are used in many Natural Language Processing (NLP) tasks and have been recently adapted to vision problems \cite{self_alone}. 

\subsection{CNN Architecture}
\begin{figure}
\centering
	\caption{\small{\textbf{CNN model.}}}
\includegraphics[width=0.7\linewidth]{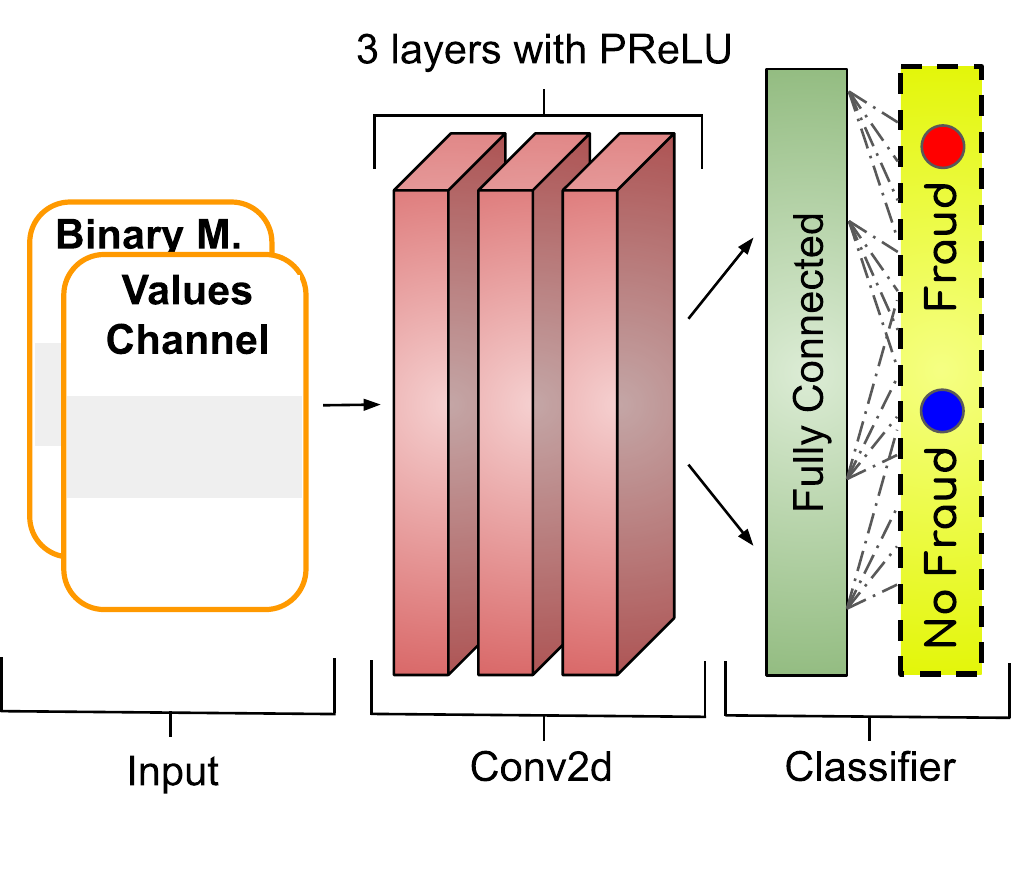}
\label{cnn_model}
\end{figure}

Our CNN model has $3$ layers of 2D convolutional operators with kernel size of 3: First layer has $2$ channels as input and $64$ as outputs; The second layer outputs $64$ channels with a non-linear activation PReLU \cite{He2015}; The third and final convolutional layer outputs $32$ channels over a dilated kernel with a stride factor of $2$ which is followed by PReLU activation function. All convolutional layers have kernel size $3$. The convolutional output is flattened and connected to a fully connected layer, Figure \ref{cnn_model} summarizes the model.

Dilation is a practice to increase the receptive view using sparse filters \cite{dilation}. The convolution itself is modified to use the filter parameters in a sparse way as it skips a fixed number of features along both dimensions at regular intervals, albeit the sparsity, dilated convolutions do not lose resolution. The stride or sub-sampling factor as mentioned in \cite{ClassifierTodense} is the step of the convolution used to reduce the overlap of receptive fields and spatial dimensions which can be seen as an alternative to pooling layers.  

\subsection{Multi-heads Attention Architecture}
Attention mechanisms have shown great ability to solve many kind of problems, ranging from NLP tasks \cite{devlin2018bert} to computer vision \cite{GoogleAtt} and tabular data \cite{arik2019tabnet}. Inspired by the recent advances we propose a novel Neural Network that takes advantage of both attention mechanisms and convolutional layers that are concatenated and unified through a convolution of kernel size $1$. We start by describing the inner works of the convolutional part.

\textbf{Convolutional Layer}: Our convolutional layer is composed of two parts, one will perform standard convolutions over the inputs, while the other part applies a convolution with dilation factor of $2$, both layers utilizes a kernel size of $3$, the results are concatenated to form a single output.

\textbf{Attention Mechanism}: Our attention mechanism differs from standard approaches by looking at the channels of the input as the heads and mapping them to another set of attention heads, that is, given an input of shape $(C, L, D)$ we first transpose the first two dimensions and flatten it into a matrix of shape $X\in\mathbb{R}^{L \times CD}$, let $W_{q}, W_{k}, W_ {v}\in\mathbb{R}^{CD \times \overline{C}D}$ be learnable linear transformations, where $C$ is the number of channels or heads coming in, $L$ is the size of the sequence, $D$ is the dimension of every element in the sequence and $\overline{C}$ is the number of output heads or channels, we start by computing $O_{q,k,v} = X W_{q,k,v}$, $O_{q,k,v} \in \mathbb{R}^{L \times \overline{C}D}$. Second we map $O_{q,k,v}$ back to a tri-dimensional shape by unflatenning and transposing so that $\overline{O}_{q,k,v} \in \mathbb{R}^{\overline{C} \times L \times D}$, finally we compute the output of the attention layer as follows:
\begin{equation}
\texttt{Attn} = \texttt{Softmax} \left(\dfrac{\overline{O}_{q} ~ \overline{O}_{k}^{T}}{\sqrt{D}} \right) ~ \overline{O}_{v} \end{equation}
 
Summarizing, given an input $X$ we perform the following mapping:

\begin{equation}
f: X \in \mathbb{R}^{C \times L \times D} \rightarrow \texttt{Attn} \in \mathbb{R}^{\overline{C} \times L \times D}    
\end{equation}

This allows for consistency of the output shape between the attention and convolutional layers. 

\textbf{Unification}: After the input is processed both by the attention and convolutional layers we concatenate the results into a single matrix and unify it through a convolution of kernel size $1$ followed by Layer Norm and PReLU activation function. We called this a \texttt{Hybrid Multi-Head Attention/Dilated Convolution Layer}.

\textbf{Classifier}: Finally the output of a sequence of these hybrid layers is flattened and fed to a linear feedforward neural network that will predict the input class. 

\begin{figure}
\centering
	\caption{\small{\textbf{Hybrid Multi-Head Attention/Dilated Convolution.}}}
\includegraphics[width=1.\linewidth]{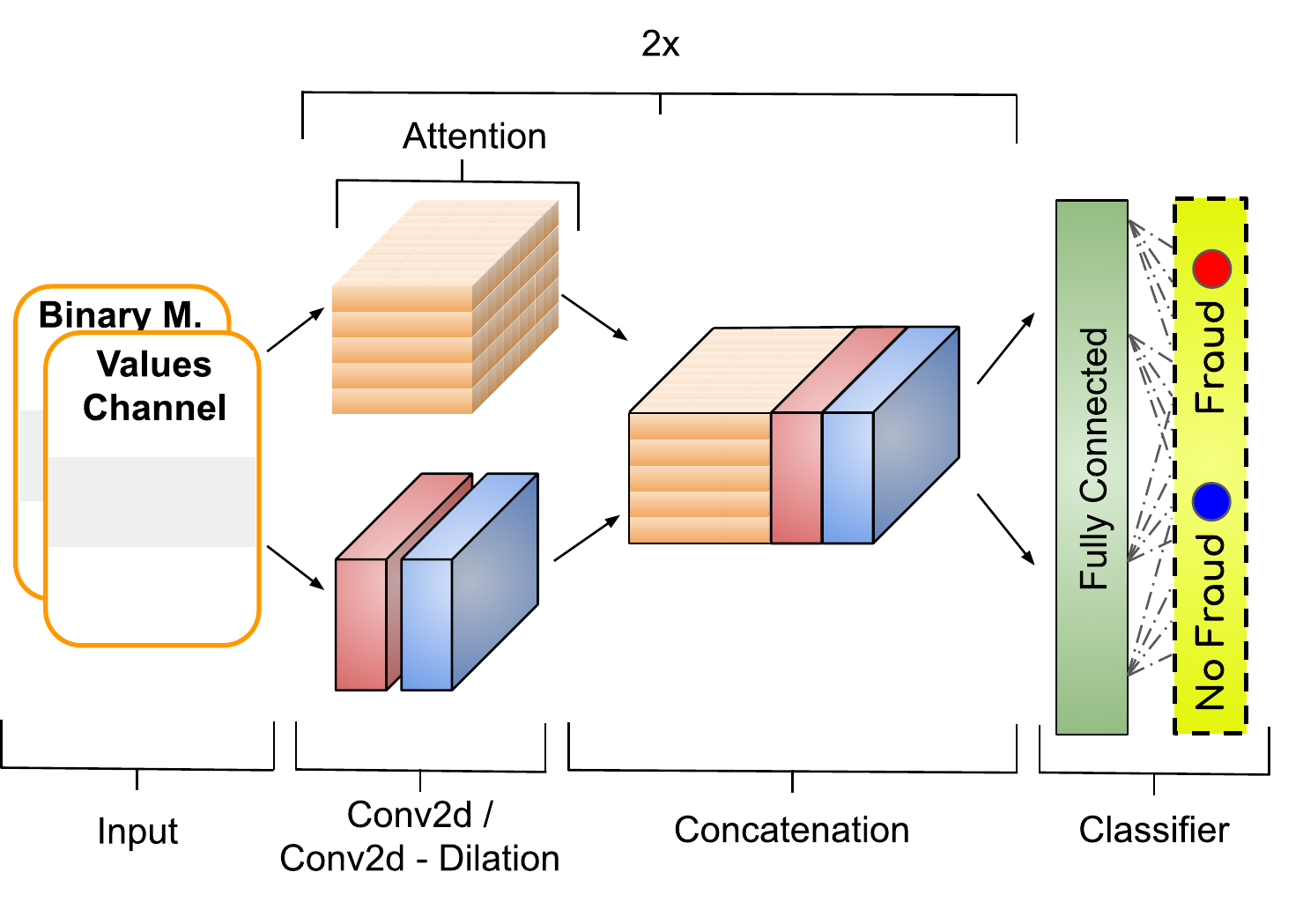}
\label{att_model}
\end{figure}

Our final architecture is composed of two hybrid layers, where the first has $C = 2$ heads and outputs $\overline{C} = 16$ heads while the convolutional part receives a $2$ channel 2D input and outputs a $32$ channel matrix of the same size, the unification is fed to a second hybrid layer with the same dimensions, lastly a one layer dense neural network with PReLU as activation function and $1024$ neurons on its hidden layer classifies the input. Figure \ref{att_model} shows the model.

\subsection{Metrics}

In this work we evaluate our models with AUC that represents the data separability degree and the  ROC curve which depicts the probability curve created by plotting the rate of true positives versus the rate of false positives. The AUC is the area under this curve that summarizes the ROC curve in a single value. 

We also evaluate the performance on the F$_1$ score that combines precision and recall in order to bring a unique number that indicates the general quality of the model. Besides these metrics we use the Mean Average Precision (MAP) \cite{MAP} to measure the effectiveness of information retrieval. To evaluate the MAP we first ordered the \textit{true labels} by the predicted probabilities and consider a subset of top $K$ probabilities given by the following equation:

\begin{equation}
\text{MAP}@K = \dfrac{1}{\sum_{i=1}^{K} r_i}\sum_{i=1}^{K} r_i\left(\dfrac{\sum_{j=1}^{i}r_j}{i}\right),    
\end{equation}
where the $r_i$ is the true label of the $ith$ consumer, $r_i=1$ if is a thief and $0$ otherwise. For the loss function we decided to use the cross entropy which is a classic practice for classification problems.

\section{Data}
\label{section_data}
The SGCC data presents the daily consumption of $42372$ consumer units with a total time window ranging from January $2014$ to October $2016$, corresponding to approximately $147$ weeks. The data is divided into thieves and normal electrical consumers, where the first compose $8.55\%$ of the total. This data does not show the date when the fraud occurs. We tested data reshape 2D on a monthly and weekly basis, we decided to use a weekly period, as we noticed a more correlation between thieves and normal electricity customers. 

Due to the granularity of the data, it is common to have a significant number cases of missing values and there are approximately 25\% of them. Our propose to handle the missing data was presented in section \ref{section_missing_data}. The dataset description is showed in the Table \ref{data_description}. The same dataset was analyzed in \cite{Zheng2018}, where the authors used an Wide and Deep architecture \cite{Cheng2016}, more details about this study is described in section \ref{section_baselines}.

\begin{table}
	\caption{\small{\textbf{Dataset Description}}}

\begin{center}

\resizebox{0.5\textwidth}{!}{%

\begin{tabular}{lll}
\midrule
\midrule
Description & Value \\ 
\midrule
Time window                    & 2014/01/01 – 2016/10/31      \\
Normal electricity customers   & 38 757 approx. 91.5\%       \\
Electricity thieves            & 3 615  $\,\;$approx. 8.55\% \\  
Total customers                & 42 372                       \\
Missing data cases             & approx. 25\% \\ 
\midrule
\midrule
\end{tabular}

}
\label{data_description}
\end{center}
\end{table}

\subsection{Data Preprocessing}
\begin{figure}
\centering
	\caption{\small{\textbf{Correlation Matrix. Top: Normal Electricity Customers, Bottom: Thieves.}}}
\includegraphics[width=0.8\linewidth]{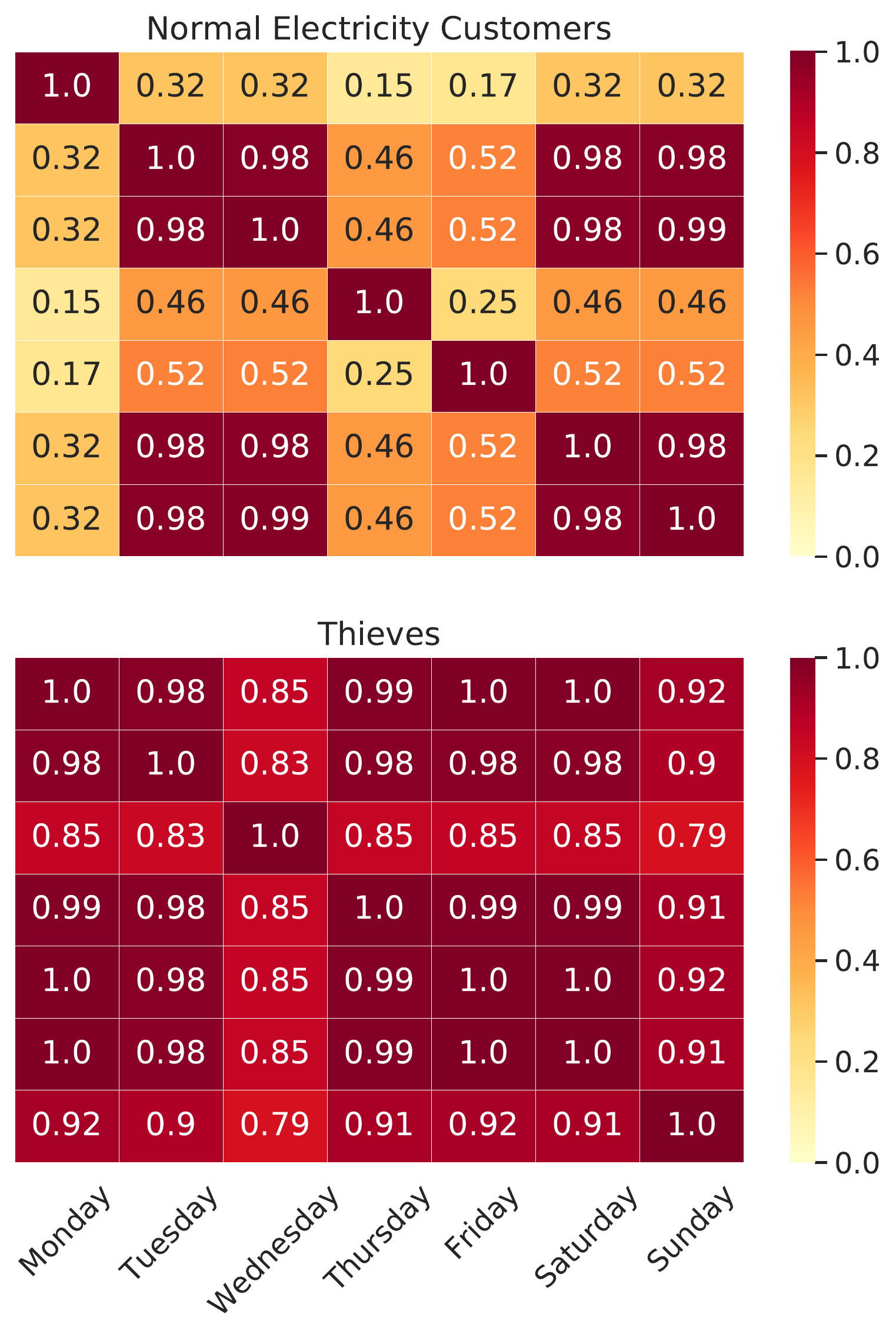}
\label{matriz-corr}
\end{figure}

Data processing is a key element that determine the success or failure in many deep learning models. In our analysis the realistic SGCC data has some particular features, including a significant number of missing data, a long tail distribution which produces strong skewness and kurtosis. The missing data is discussed in section \ref{section_missing_data}. For the atypical data, or \textit{outliers}, we noticed that most of the cases occur in the normal electricity costumers and we did not remove these cases to avoid losing useful information. Prior to the normalization of the data, we studied the dataset as a time series due to the fact that there is only one variable performed at uniform intervals. To evaluate possible correlations and periodicity, two experiments were conducted: \texttt{(I)} we accumulated the electricity consumption over the $7$ days of the week (from Monday to Sunday) and constructed a correlation matrix between the days of the week for thieves and normal electricity customers, as illustrated in Figure \ref{matriz-corr}.

\begin{figure}
\centering
	\caption{\small{\textbf{Autocorrelation of Electricity Consumption. Top: Normal Electricity Customers, Bottom: Thieves.}}}
\includegraphics[width=0.8\linewidth]{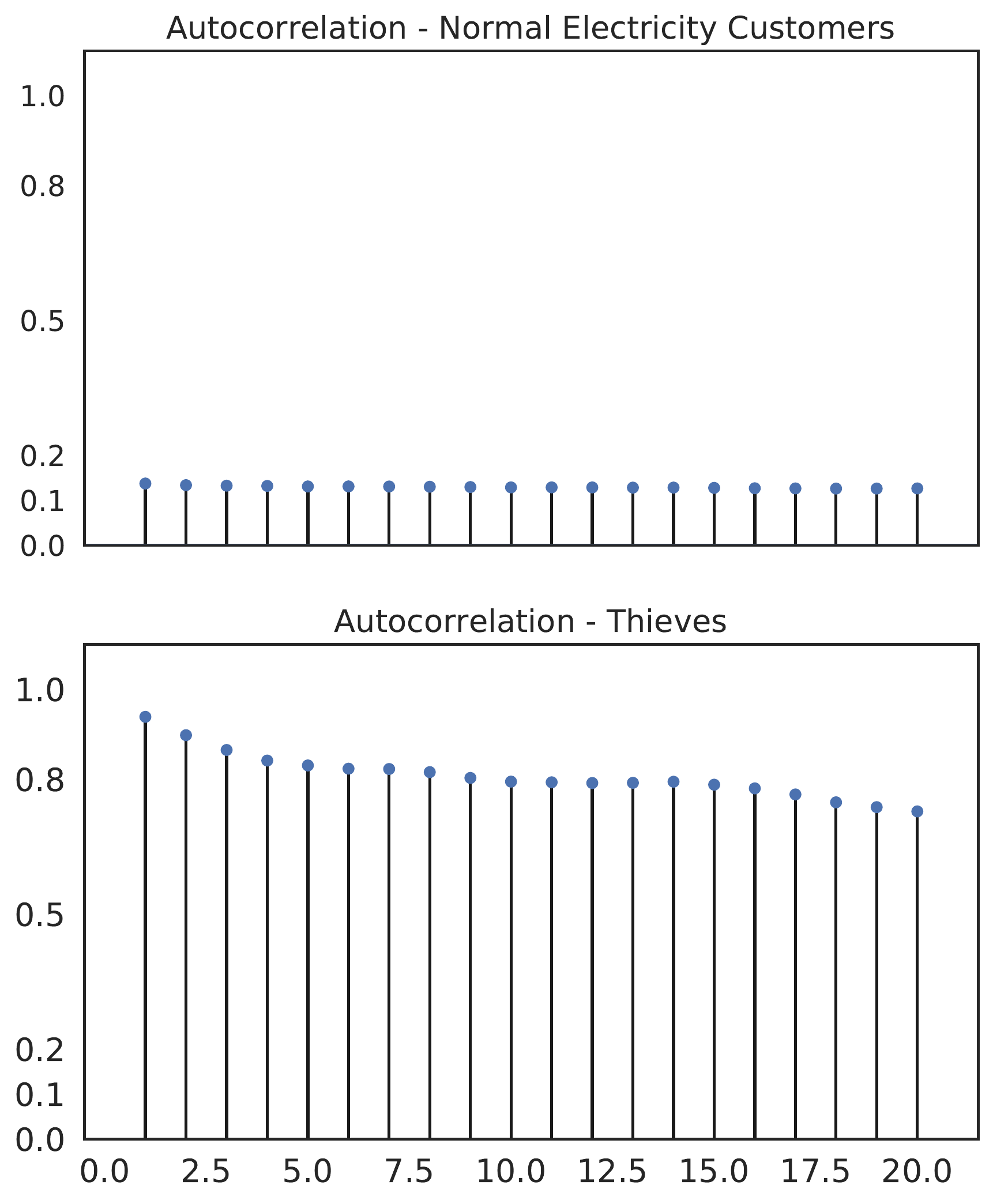}
\label{autocorr}
\end{figure}

\noindent\texttt{(II)} In order to find periodicity and pattern recognition between classes we use the autocorrelation function, which provides the correlation of a time series with its own lagged values, Figure \ref{autocorr}. The $x$ axis indicates the interval $t-$time being considered, where $t=20$ meaning a lag of $20$ intervals; $y$ axis is the autocorrelation score and $y=1$ is the highest possible score. Top: normal electrical customers, Bottom: thieves.

The analysis from Figures \ref{matriz-corr} and \ref{autocorr} shows some difference between thieves and normal electricity customers. In particular, the greater correlation observed between days of the week for the thieves suggests that this feature could be exploited to improve model performance, in another words, the thieves have similar behaviour. 

\begin{figure}[H]
\centering
	\caption{\small{\textbf{Electrical Consumption Data from 100 samples (in blue). Top: Raw data; Bottom: Data processed by Quantile transformation.}}}
\includegraphics[width=\linewidth]{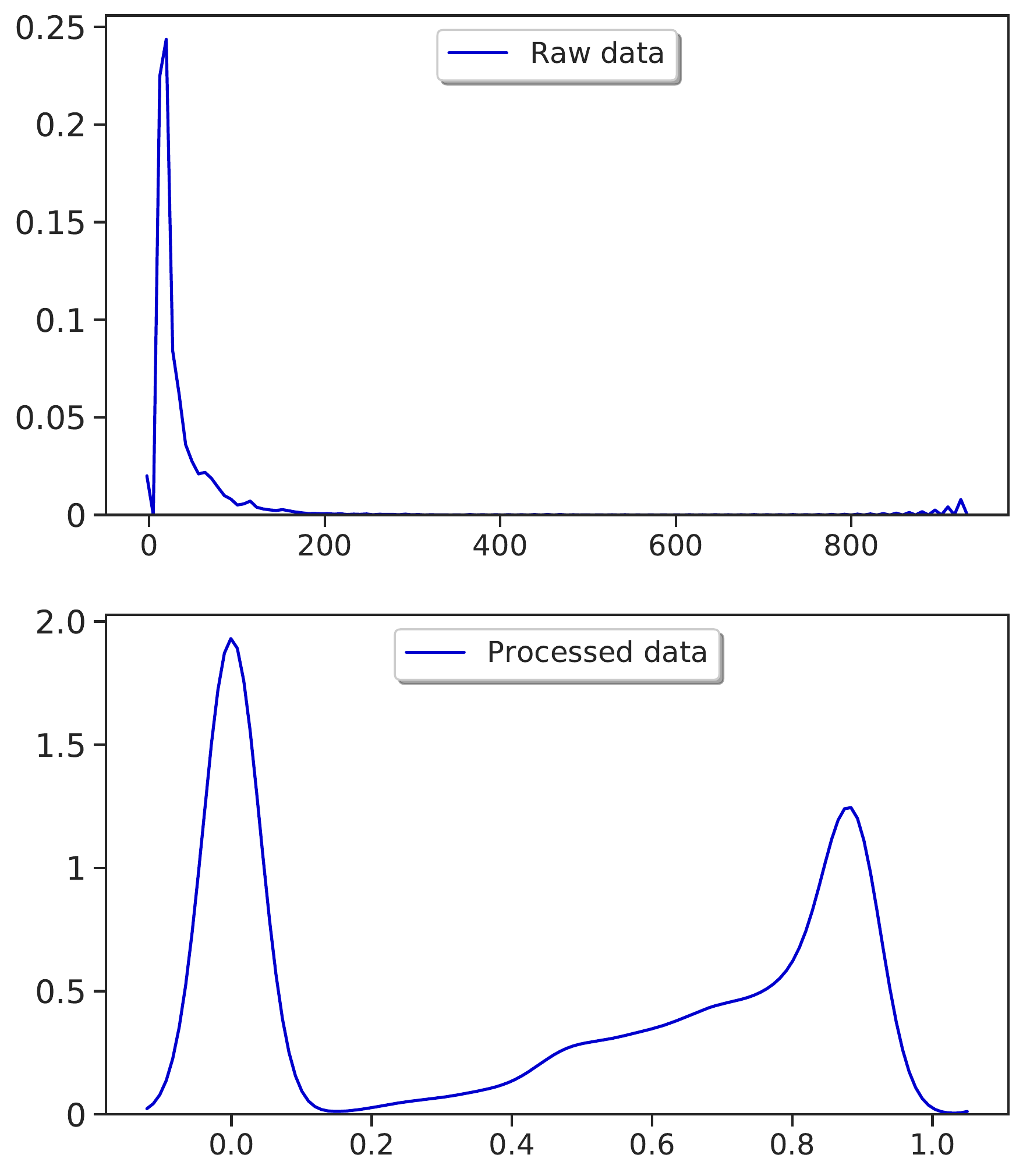}
\label{fig_kl}
\end{figure}

The SGCC data has a phenomenon called heteroscedasticity (non-constant variability) \cite{hetero}, which causes the resulting distribution to be asymmetric positive or Leptokurtic \cite{leptokurtic}, i.e., there is great variability on the right side of the distribution which creates a long tail, as shown in Figure \ref{fig_kl}-Top. This asymmetry can lead to spurious interactions in the deep learning model due to non-constant variations. To deal with this asymmetry distribution we perform a Quantile uniform normalization provide by \cite{scikit-learn}. The Quantile uniform transformation is a non-linear function which is applied on each feature data independently. This normalization spreads out the most frequent values between $(0,1)$. First, the Quantile map the original values to estimate the cumulative distribution, then these values are spread out into numbers of quantiles. In our approach we use $10$ quantiles. A distribution of the data processed is shown in Figure \ref{fig_kl} on the Bottom. One problem that Quantile transform has is the the number of data required to performed the transformation. As a rule of thumb, to create $m$ quantiles, a minimum of $10\times m$ samples is required. 

In addition to processing Quantile, we also tested a Yeo-Johnson power transform \cite{yj}, but the transformed values were between $[0, 12]$ and with Quantile between $[0,1]$. We also verified the Kullback-Leibler Divergence ($D_{KL}$) \cite{kullback} to a uniform distribution is minimized. $D_{KL}$ is a practice of measuring the matching between two distributions, given by the formula: 

\begin{equation}
D_{KL}(p || q) = \sum_{j=i}^{N}p(x_j)\log \left( \dfrac{p(x_j)}{q(x_j)} \right),     
\end{equation}
where $q$ is the distribution of the data transformed by Quantile and $p$ is the ground truth, in our case a uniform distribution and we are interested in matching $q$ to $p$. A lower $D_ {KL}$ value means a better $p$ and $q$ matched. The Table \ref{tab_processing} shows the $D_ {KL}$ values before and after Quantile transformation.

The processed dataset has less Kurtosis and Skewness, which brings stationarity to the data by Kwiatkowski, Phillips, Schmidt and Shin (KPSS) \cite{kpss} test with $\alpha$ level equals $5\%$. Namely the data variance, mean and covariance has more stationary behavior and its statistical properties do not change over time in the columns where the KPSS test is \texttt{True}, Table \ref{tab_processing}.  

\begin{table}
	\caption{\small{\textbf{Processing data}}}
\begin{center}
\label{tab_processing}

\resizebox{0.5\textwidth}{!}{%

\begin{tabular}{lcccc}
\midrule
\midrule
Property  && \textbf{Raw data}  && \textbf{Processed data} \\ 
\midrule
Min        && 0.00                     && 0.00    \\
Max        && 800003.31                && 1.00    \\
Mean       && 6.87                     && 0.40    \\
Std        && 236.14                   && 0.35    \\
Skewness   && 2551.62                  && -0.01   \\
Kurtosis   && 7170709.11               && -1.67   \\
$D_{KL}$         && 15121.81                 && 57.15   \\
KPSS test  && \textit{False:} 1016 / \textit{True:} 19   && \textit{False:} 581 / \textit{True:} 454\\
\midrule
\midrule
\end{tabular}
}
\label{tab:processed}
\end{center}
\end{table}

\section{Experiments}
\label{sec:experiments}
In this section we describe the experiments performed in this work. In addition to the two models developed, we also compared our attention model with the Attention Augmented Convolutional Network \cite{GoogleAtt}. To evaluate the proposed modification for the missing data described in section \ref{section_missing_data}, we also performed an experiment with and without a Binary Mask. All training sessions were performed with different train percentages splits and with stratified k-fold. 

\subsubsection{Binary Mask Experiment}
Using stratified k-fold with the \texttt{Hybrid Multi-head Attention Dilation Convolutional} model and training split $=80\%$ we evaluated the percentage difference of the data with Binary Mask and without. When there's Non-Binary Mask, all missing data was filled with $0$ value, Table \ref{experiment_table} presents results of this experiment where the column name \textit{Only Quantile} refers to Non-Bynary Mask.

\subsubsection{Attention Augmented Convolution Network}
We implemented the Attention Augmentation Convolutional Network algorithm proposed in \cite{GoogleAtt}. Which is a self-attention algorithm developed for two-dimensional tasks as an alternative to CNN networks. The authors combine features extracted from the convolutional layers with self-attention through concatenation. The experiment was performed with stratified k-fold in different train splits size. Table \ref{results} shows the results.

\subsection{Baselines}
\label{section_baselines}
Detection of electrical fraud with granular data using Deep Learning techniques are still rare to be found in the literature. The dataset on which this work was developed is a real data, which makes it even rarer. To compare our model with other approaches, we will use \cite{Zheng2018} that made the dataset available. These authors developed a study with Wide and Deep technique \cite{wideAndDeep}. The Wide component try to memorize the global knowledge and the CNN layers capture features of electricity consumption data. These two components associated resulted in a good performance with an AUC metric up to $0.79\%$ and $MAP@100$ above $0.96$.

\subsubsection{Dataset preprocessed with Missing Values Interpolated}
Our aim in this experiment is to conduct:
\begin{itemize}
    \item The Quantile transformation contributed positively to our preprocessing data proposal
    \item The Hybrid Multi-Head Attention/Dilated Convolution outperformed the Wide and Deep model \cite{Cheng2016} in the same data. 
\end{itemize}
For this, we preprocessed the SGCC dataset with the equations $1, 2$ and $3$ as in Zheng et al.~\cite{Zheng2018} and trained our model in the split $80\%$ with stratified k-fold. Results are presented in Table \ref{experiment_table}, column name \texttt{Interpolated Missing Values}. With the same dataset configuration as our baseline, we improve all the metric scores and the results presented are the average values for all folds at the same epoch. To show the Quantile transformation is efficient, we need to compare the results obtained in Table \ref{experiment_table} between the columns name \texttt{Only Quantile} and \texttt{Interpolated Missing Values}.

\begin{table}[H]
	\caption{\small{\textbf{Binary Mask Experiment: all columns was trained with Hybrid Multi-Head Attention/Dilated Convolution with train split = 80\%.}}}
\begin{center}
\resizebox{0.5\textwidth}{!}{%

\begin{tabular}{lcccc}
\midrule
\midrule
                & \textbf{Interpolated}   &   \textbf{Only}   & \textbf{Quantile +}\\
\textbf{Metric} & \textbf{Missing Values} & \textbf{Quantile} &\textbf{Binary Mask} \\

\midrule
AUC             & 0.840                   & 0.889                & \textbf{0.925} \\
F$_1$ score     & 0.365                   & 0.504                & \textbf{0.606} \\
MAP@100         & 0.960                   & 0.972                & \textbf{0.992} \\
MAP@200         & 0.941                   & 0.961                & \textbf{0.972}  \\
\midrule
\midrule
\end{tabular}
}
\label{experiment_table}
\end{center}
\end{table}

\subsection{Results and Discussion}
\begin{figure}[H]
\centering
	\caption{\small{\textbf{Metrics by Epochs, train = $80\%$.}}}
\includegraphics[width=\linewidth]{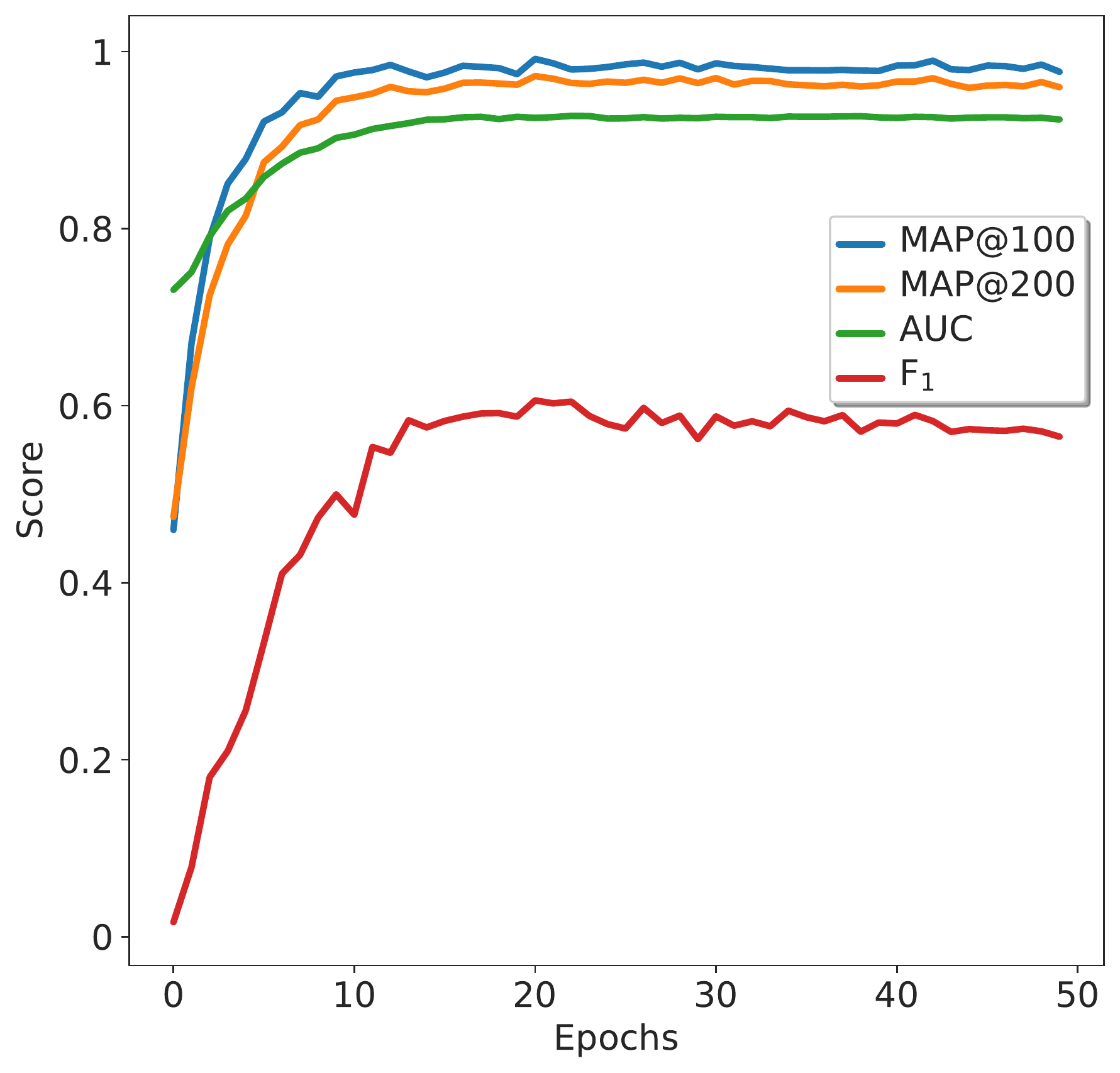}
\label{metrics_by_epochs}
\end{figure}

Table \ref{results} presents the main results of the models developed in this work. The three train splits of $50\%, 75\%$ and $80\%$ were tested with stratified k-fold. The Hybrid Multi-Head Attention Dilation/Convolution significantly outperformed the baseline. Moreover, the score obtained with our two models and with the Attention Augmented Convolutional Newtwork shows that the Quantile transformation brings a significant improvement to the data preprocessing. The attention mechanism produces a notable increase in the F$_1$ score. Another distinguished behaviour is the much faster convergence of the attention model compared with the CNN model. In our tests the CNN needed approximately $100$ epochs to converge while the Hybrid Attention converged with approximately $20$ epochs. Figure \ref{metrics_by_epochs} presents the evolution of the scores as a function of the epoch for the Hybrid Multi-Head Attention Dilation/Convolution.

\begin{table}[H]
	\caption{\small{\textbf{Main Results}}}
\begin{center}
\resizebox{0.5\textwidth}{!}{%

\begin{tabular}{llccc}
\midrule
\midrule
\textbf{Model} & \textbf{Metric} & train = 50\% & train = 75\% & train = 80\% \\ 
\midrule
 
\textbf{\texttt{Conv.}}      & AUC         & 0.898 & 0.920  &  0.922 \\
\textbf{\texttt{Neural}}     & F$_1$ score & 0.477 & 0.508  &  0.530 \\
\textbf{\texttt{Network}}    & MAP@100     & 0.977 & 0.978  &  0.979 \\
                             & MAP@200     & 0.969 & 0.970  &  \textbf{0.976} \\
\midrule
\textbf{\texttt{Hybrid}}     & AUC         & \textbf{0.903} & \textbf{0.926}  &  \textbf{0.925} \\
\textbf{\texttt{Multi-Head}} & F$_1$ score & \textbf{0.553} & \textbf{0.583}  &  \textbf{0.606} \\
\textbf{\texttt{Attention}}  & MAP@100     & \textbf{0.996} & \textbf{0.988}  &  \textbf{0.992} \\
\textbf{\texttt{Dil. Conv.}} & MAP@200     & \textbf{0.981} & \textbf{0.971}           &          0.972 \\
\midrule
 
\textbf{\texttt{Attention}}  & AUC         & 0.881 & 0.902  &  0.911   \\
\textbf{\texttt{Augmented}}  & F$_1$ score & 0.503 & 0.543  &  0.551  \\
\textbf{\texttt{Conv.}}      & MAP@100     & 0.956 & 0.969  &  0.969   \\
\textbf{\texttt{Network}}    & MAP@200     & 0.948 & 0.956  &  0.952   \\
\midrule
\midrule
\end{tabular}
}
\label{results}
\end{center}
\end{table}

\begin{figure}
\centering
	\caption{\small{\textbf{Threshold Analysis}}}
\includegraphics[width=\linewidth]{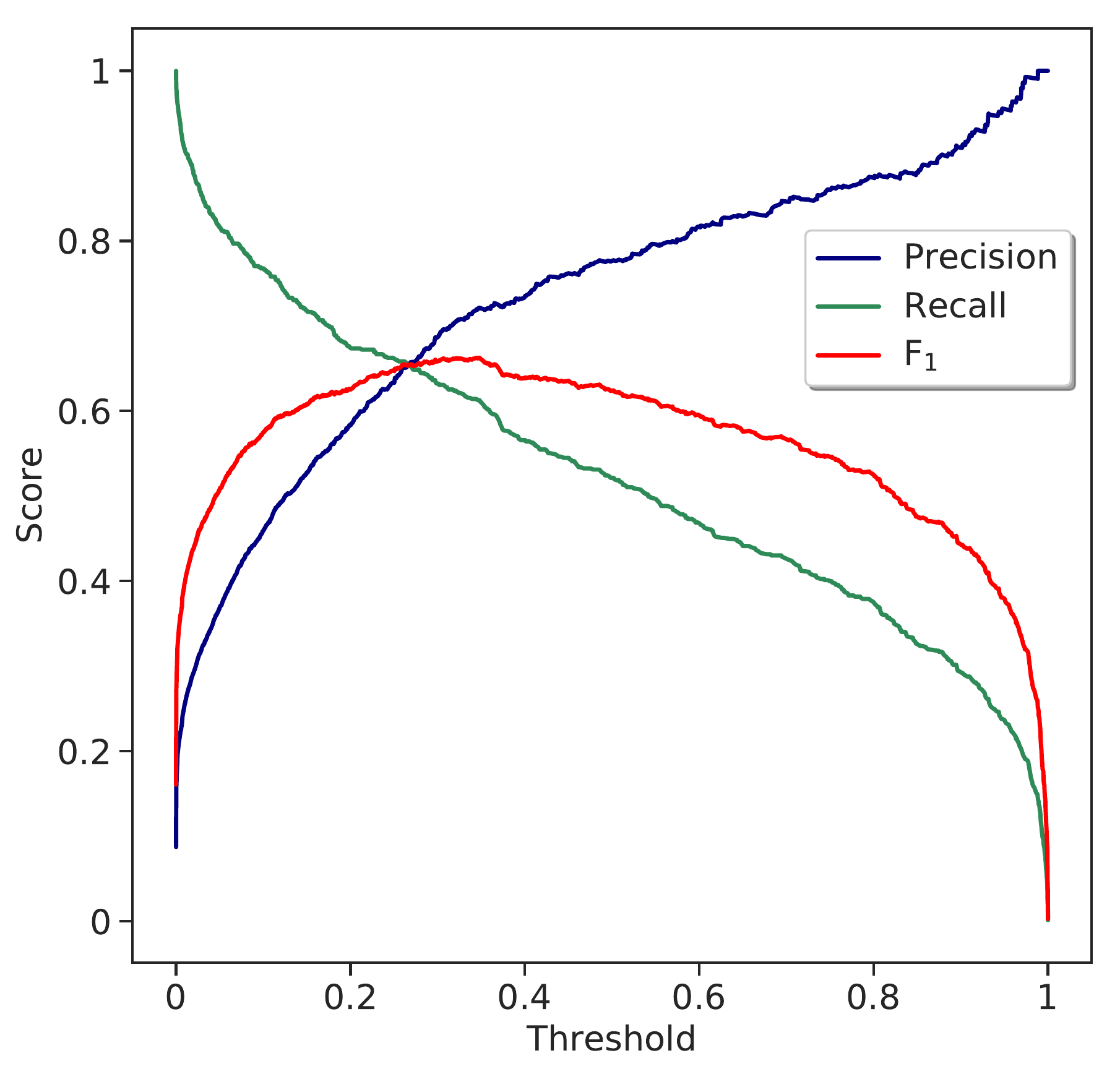}
\label{thrs_fig}
\end{figure}

\begin{figure}
\centering
	\caption{\small{\textbf{Confusion Matrix from one fold in train = 80\%}}}
\includegraphics[width=\linewidth]{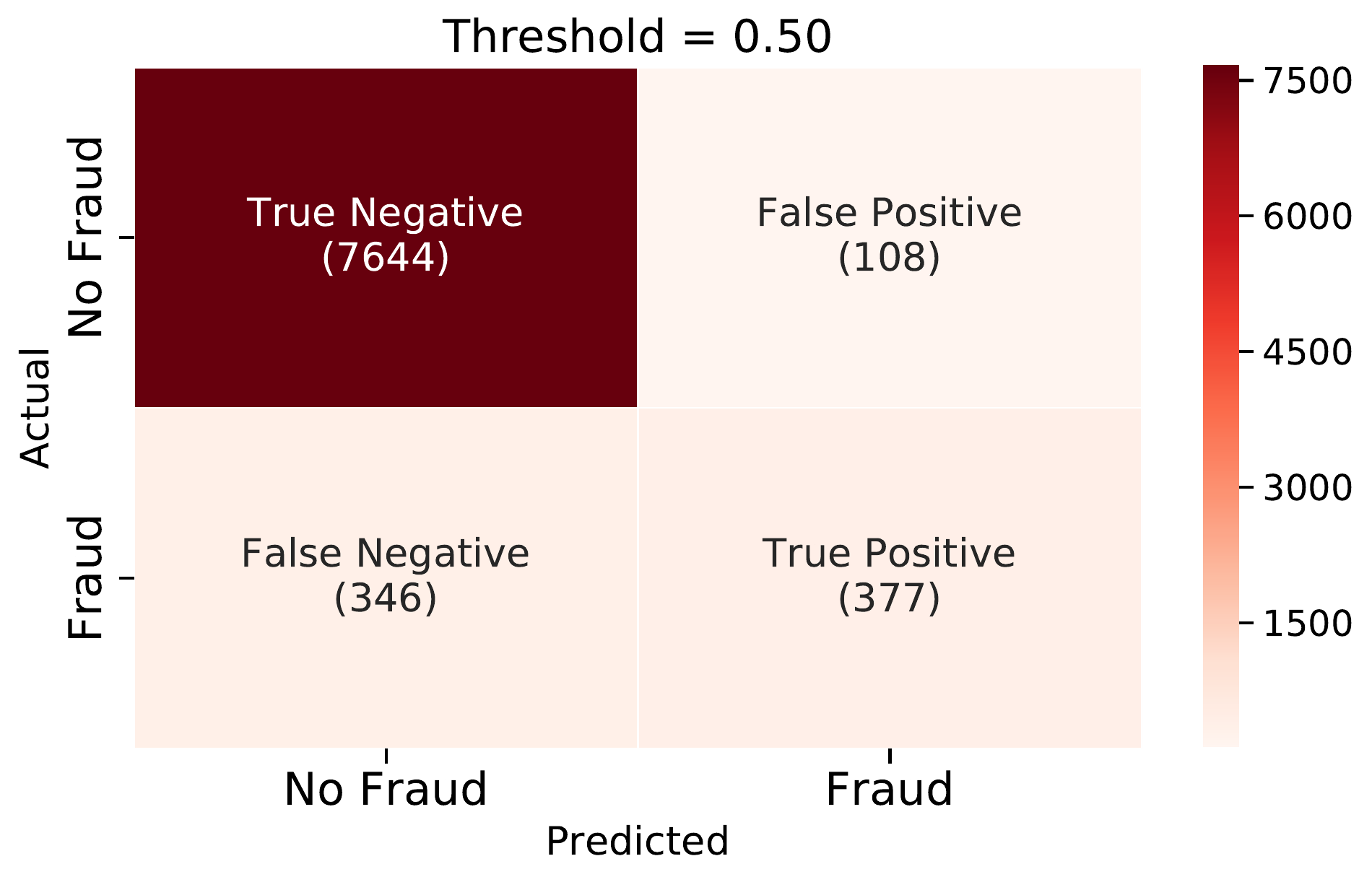}
\label{confusion}
\end{figure}

With respect the time spent during the training and inference the Table \ref{time_table} shows the average time spent for $1$ epoch in $5$ folds in the training and total time needed to infer the valid data which is $20$\% of the dataset. The results achieved enable the establishment of protocols for suspected cases inspection with high assertiveness. However, it is necessary to note that the choice of the threshold is an important point for decision making. Our model has an optimal threshold of $0.27$, as shown in Figure \ref{thrs_fig}, which produces a F$_1$ score of $0.65$. Note that when a $0.50$ threshold is used there is a trade-off between Precision and Recall. In other words, if Precision is prioritized, we must choose a threshold greater than $0.27$. The Table \ref{results} and the confusion matrix in Figure \ref{confusion} correspond to threshold $0.50$.

\begin{table}[!htbp]
	\caption{\small{\textbf{Training time, on Tesla V100 GPU Hardware - with train split = 80\%}}}
\begin{center}
\resizebox{0.5\textwidth}{!}{%

\begin{tabular}{lcccc}
\midrule
\midrule
\textbf{Model} & \textbf{Training time}   &   \textbf{Inf. time}   & \textbf{\# Params.}\\
\midrule
CNN             & 2min 27seg           & 32seg                & 3Mi \\
Hybrid Attn.    & 3min 16seg           & 37seg                & 51Mi \\
Attn. Augmented & 2min 40seg           & 20seg                & 17Mi \\
\midrule
\midrule
\end{tabular}
}
\label{time_table}
\end{center}
\end{table}

\section{Conclusion}
\label{sec:conclusion}
In this paper, we introduced a Hybrid multi-head self-attention dilated convolution method for electricity theft detection with realistic imbalanced data. We apply three innovations to improve upon the previous baseline work:
\begin{enumerate}
    \item A Quantile normalization of the dataset;
    \item The introduction of a second channel to the input called Binary Mask;
    \item A novel model of multi-head self-attention.
\end{enumerate}
Another key element is the time series data reshape in 2D format introduced by \cite{Zheng2018,li_electricity_2019} allowing to treat the consumer sample as an image and to use CNNs. Our attention model overperformed the CNN model developed up to $5$ points of F$_1$ and converged in $20$ epochs, approximately $1$hour and $9$min compared with $100$ epochs in CNN, approximately $4$hours and $8$min. 

The model presented in \cite{GoogleAtt} was the inspiration for our attention model. The unification step that combines the outputs from the attention, normal and dilated convolution, allowing that information from different spatial sizes and sources be merged, is the core of our model's architecture. The characteristics of our model do not emerge from the used data, that said, problems on computer vision, for instance, could also be solved by it.

Due to the high number of missing values in the data (approx. 25\%). Classic attempts  to reconstruct these values can bring a significant bias resulting in poor solutions. With the addition of the Binary Mask we improved the F$_1$ score em approximately $10$ points to the best of our knowledge this is the first time that the a Binary Mask was introduced as input channel into a CNN for dealing with missing data. Deep learning solutions in electricity theft detection are rare in the literature. To incentive the research in this field we are providing the code in a repository of GitHub \scriptsize{\url{https://github.com/neuralmind-ai/electricity-theft-detection-with-self-attention}} \normalsize{ and the dataset can be found at another repository} \scriptsize{\url{https://github.com/henryRDlab/ElectricityTheftDetection/}} \normalsize{.
The results obtained in this study demonstrate that still exist space for advances into the results obtained by Deep Learning techniques applied to electricity theft detection in smart real metered data.}

\subsection{Future Work}
The insights produced and experience gained from this work will be used in future experiments involving energy such as energy consumption forecasting and fraud detection in the context of another AMI framework, where data will be available at almost real time with higher sampling rate.

\subsection{Acknowledgments}
This work is funded by ENEL in ANEEL R\&D Program PD\_06072\_06 61/2018.
Roberto Lotufo thanks CNPQ's support through the research project PQ2018, process number $310828 / 2018- 0$.


\end{document}